\documentclass[11pt]{article}

\usepackage{svg}
\usepackage{multirow}
\usepackage[most]{tcolorbox}
\usepackage{courier}
\usepackage{float}
\usepackage{microtype} 
\usepackage{booktabs}  
\usepackage{url}  
\usepackage{placeins}
\usepackage{tabularx}
\usepackage{xcolor}
\usepackage{amsmath}
\usepackage{amsthm}
\usepackage{amssymb}
\usepackage{pifont}
\newcommand{\cmark}{\ding{51}}%
\newcommand{\xmark}{\ding{55}}%
\usepackage{titlesec}

\titlespacing*{\paragraph}
{0pt}      
{0.4em}    
{0em}    

\usepackage[final]{automl}

\newtcblisting{promptbox}[1]{
  enhanced,
  breakable,
  colback=gray!5,
  colframe=black!70,
  title=#1,
  fonttitle=\bfseries,
  boxrule=0.5pt,
  arc=2pt,
  left=6pt,
  right=6pt,
  top=6pt,
  bottom=6pt,
  listing only,
  listing options={
    basicstyle=\ttfamily\footnotesize,
    breaklines=true,
    columns=fullflexible,
    keepspaces=true,
    showstringspaces=false
  }
}

\usepackage[numbers]{natbib}
\title{
Large Language Models for Automated Cross-Domain Machine Learning Task Type Identification: A Benchmark Dataset and Evaluation}
\author[1]{\nameemail{Petros Tsialis}{petros.tsialis@hs-aalen.de}}

\author[2]{\nameemail{Steffen Limmer}{steffen.limmer@honda-ri.de}}

\author[2]{\nameemail{Tobias Rodemann}{Tobias.Rodemann@honda-ri.de}}

\author[1]{\nameemail{Martin Heckmann}{martin.Heckmann@hs-aalen.de}}

\affil[1]{University of Applied Sciences Aalen}

\affil[2]{Honda Research Institute Europe}
\hypersetup{
  pdflang=en,
  pdfauthor={},
  pdftitle={},
  pdfsubject={},
  pdfkeywords={}
}

\begin{document}
\maketitle
\begin{abstract}
Machine learning task type identification is essential for constructing valid ML pipelines, yet in practice it is typically specified manually.  
We investigate whether large language models (LLMs) can infer both the data domain and the downstream prediction task directly from dataset-level information when only the target feature is provided by the user. 
Together with our LLM-based system we also release an annotated benchmark comprising 625 public tabular and time series datasets. 
We evaluate the proposed approach in three settings: (i) tabular datasets in comparison with established AutoML heuristics, (ii) cross-domain evaluation across tabular and time series datasets, and (iii) a practical deployment scenario using smaller local models. 
The results show consistent advantages for LLM-based task type identification, with increasing difficulty in heterogeneous and resource-constrained settings. 
LLM-based approaches outperform AutoGluon in the tabular setting, reaching 0.98 F1 macro compared to 0.93. In the cross-domain setting, the best model achieves 0.90 F1 macro, while smaller locally deployable models reach 0.75, indicating a trade-off between deployment feasibility and accuracy. 

\end{abstract}

\vspace{-0.9em}
\section{Introduction}
\label{sec:introduction}


Identifying the correct machine learning (ML) task for a dataset is a fundamental prerequisite for building effective ML pipelines, since preprocessing, model selection, and evaluation depend on the task formulation. However, most Automated Machine Learning (AutoML) systems assume that the downstream task, such as classification, regression, or time series forecasting, is already known. In practice, this step is often performed manually by data scientists or domain experts \cite{karmaker_santu_automl_2021}, which remains challenging in industrial settings where datasets are heterogeneous, poorly documented, and frequently handled by non-expert stakeholders \cite{lewis_characterizing_2021}. Although AutoML frameworks have been applied across tabular data \cite{mueller_faster_2020}, time series \cite{wang_towards_2022,wang_automated_2019,baratchi_automated_2024}, and natural language processing \cite{shi_multimodal_2021}, task type identification is often either not addressed explicitly \cite{feurer_auto-sklearn_2022,hutter_tpot_2019,jin_autokeras_2023,wang_flaml_2021,ali_pycaret_2020,noauthor_lightwood_2026,dong_gama_2021} or implemented through heuristic and rule-based methods \cite{erickson_autogluon-tabular_2020-1,ledell_h2o_2020,mohr_naive_2023}. These methods are mainly designed for downstream task identification in tabular settings and generally do not address domain identification, i.e., distinguishing the structural machine learning problem domain of a dataset, such as tabular data or time series data, across heterogeneous data modalities. 

Recent advances in large language models (LLMs) suggest a possible alternative. LLMs have shown strong capabilities in processing structured inputs \cite{sui_table_2024}, extracting semantic patterns \cite{trirat_automl-agent_2025}, and solving classification tasks \cite{agisha_ntwali_detection_2025}. Their ability to combine raw feature values, statistical summaries, and textual context makes them a promising candidate for identifying both the data domain and downstream task from dataset-level information. To study this problem systematically, we introduce an openly available benchmark for machine learning task type identification\footnote{\url{https://zenodo.org/records/21649607}} and propose an LLM-based system\footnote{\url{https://github.com/ptsialis/Machine-Learning-Task-Type-Identification}} that infers task information from structured dataset representations, target-specific statistics, and optional textual descriptions.
To systematically investigate this problem, we address the following research questions (RQs):

\vspace{-0.5em}



\begin{itemize}
    \setlength{\itemsep}{0.3em}
    \setlength{\parskip}{0pt}
    \setlength{\parsep}{0pt}

    \item \textbf{RQ1: Baseline Comparison.} How accurately can LLM-based approaches identify machine learning tasks on tabular datasets, and how do they compare to established AutoML systems?

    \item \textbf{RQ2: Cross-Domain Capability.} How accurately can LLM-based systems identify machine learning tasks across different data modalities, including tabular data and time series, when using state-of-the-art foundation models?

    \item \textbf{RQ3: Practical Deployment.} What level of task identification performance can be achieved using local or resource-constrained LLMs, and how does this performance compare to large, state-of-the-art models in realistic deployment scenarios?
\end{itemize}
\vspace{-0.9em}
\section{Related Work}
\label{sec:related}
\vspace{-0.2em}
\paragraph{LLMs}
Recent developments in large language models (LLMs) have shown that these models can effectively process and reason over structured data representations \cite{tan_struct-xenhancing_2025, sui_table_2024}. In addition, LLMs have demonstrated strong performance on classification tasks \cite{hegselmann_tabllm_2023,  aragao_practical_2025}, particularly in zero-shot and few-shot settings \cite{labrak_zero-shot_2024, chae_large_2026}. Prior work further suggests that this capability can be strengthened through appropriate prompt design, for example by using zero-shot or few-shot prompting \cite{labrak_zero-shot_2024, chae_large_2026}, encouraging explicit reasoning \cite{tan_struct-xenhancing_2025}, or enriching the prompt with dataset-specific meta-information\cite{agisha_ntwali_detection_2025,sahin_boosting_2026}.

\paragraph{LLM-Assisted AutoML}
Building on these strong capabilities of LLMs, recent research has explored LLM-based approaches for automating ML workflows~\cite{gu_large_2025}. Several of these systems include some form of task interpretation, but they differ fundamentally from our setting. AutoML-GPT~\cite{zhang_automl-gpt_2023} uses an LLM to coordinate data processing, model selection, and hyperparameter tuning, but the task is specified through user-provided instructions and structured workflow prompts. AutoML-Agent~\cite{trirat_automl-agent_2025} decomposes ML workflows into sub-tasks handled by LLM agents, but also starts from an explicit task description. MLAgentBench~\cite{huang_mlagentbench_2024} evaluates language agents on end-to-end ML experimentation tasks, where each task is defined by a textual task description as well as starter files such as code and data. Thus, the agent is required to improve or solve an already specified ML task rather than infer the task type from dataset-level information. CAAFE~\cite{hollmann_large_2023} uses LLMs to generate semantic features from dataset descriptions, where the description typically provides substantial information about the prediction problem. MLCopilot~\cite{zhang_mlcopilot_2024-1} recommends ML solutions by retrieving experience from previous ML tasks, but the new task is again represented through an explicit task specification. Thus, existing LLM-assisted AutoML systems may interpret, optimize, or operationalize ML tasks, but the downstream task is usually given by the user or available through descriptions that already state the prediction problem. In contrast, we treat task type identification as a standalone problem: only the target feature is assumed to be known, while the data domain and downstream task must be inferred from dataset-level information. The model receives feature names, a serialized dataset excerpt, and target-specific statistics, optionally complemented by dataset descriptions from the dataset creators. Importantly, these descriptions generally characterize the dataset rather than explicitly defining the underlying ML task. To systematically assess their contribution to task identification, we conduct all experiments both with and without the inclusion of these dataset descriptions.

\section{Datasets}
\label{sec:dataset}
\begin{figure}[htbp]
    \centering
    \includegraphics[width=1\textwidth]{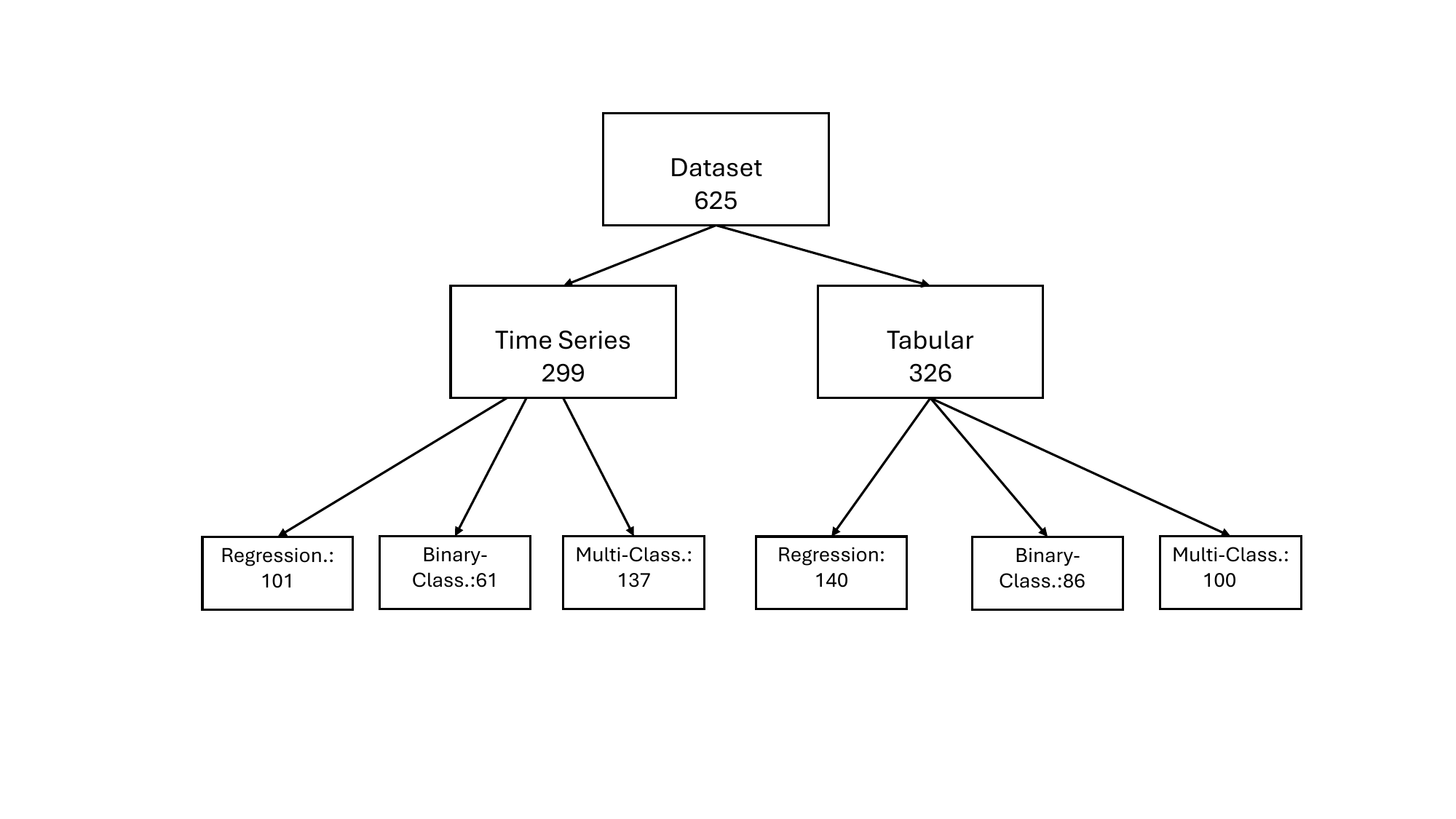}
    
    \caption{Two-level taxonomy of the benchmark dataset collection.}
    \label{fig:taxonomy}
\end{figure}

To the best of our knowledge, our collection is the first openly available benchmark specifically designed for machine learning task type identification. It enables systematic evaluation of whether models can infer both the data domain and downstream prediction task from dataset content and meta-information, while its heterogeneous sources and file formats reflect realistic data conditions. The benchmark comprises 625 public datasets from OpenML \cite{bischl_openml_2025}, the UCI Machine Learning Repository \cite{kelly_uci_nodate}, Kaggle \cite{noauthor_kaggle_2026}, and the Time Series Classification Archive \cite{middlehurst_bake_2024}. As shown in Figure~\ref{fig:taxonomy}, it follows a two-level taxonomy: datasets are first assigned to \textit{Time\_Series} or \textit{Tabular}, with 299 and 326 datasets, respectively, and then to binary classification, multiclass classification, or regression/forecasting. The time series branch contains 101 regression, 61 binary classification, and 137 multiclass classification datasets; the tabular branch contains 140 regression, 86 binary classification, and 100 multiclass classification datasets. This taxonomy supports evaluation of both domain identification and downstream task identification within and across dataset domains.

The collection consists of the original datasets and a tabular meta-information dataset which contains e.g. dataset name, data domain, downstream task, target variable, and dataset description.
A full description of the metadata fields, dataset-type details, and loading-pipeline details are provided in Appendix~\ref{app:datasets}. Since the datasets vary strongly in file format, storage structure, and representation, we developed a loading pipeline that parses different formats while preserving the original structure where possible. Extensive preprocessing into a uniform schema is avoided to better reflect realistic data conditions.

\paragraph{Tabular classification}
Tabular classification datasets consist of independent samples described by input features and an explicitly specified target feature. The target feature represents a discrete outcome. If it contains two possible classes, the dataset is treated as binary classification; if it contains more than two classes, it is treated as multiclass classification \cite{murphy_machine_2013}.

\paragraph{Tabular regression}
Tabular regression datasets also consist of independent samples described by input features and an explicitly specified target feature. In contrast to classification, the target feature represents a continuous numerical outcome. The objective is therefore to predict a real-valued response from the available input features \cite{murphy_machine_2013}.

\paragraph{Time series classification}
Time series classification datasets consist of samples whose observations are ordered over time or another sequence dimension. Each sample may be represented by a univariate or multivariate sequence, where the order of observations is part of the data representation. The target feature is categorical, and the objective is to assign each time series sample to one of the predefined classes \cite{ismail_fawaz_deep_2019,mohammadi_foumani_deep_2024}.

\paragraph{Time series forecasting/regression}
Time series forecasting and regression datasets consist of samples whose observations are ordered over time or another sequence dimension, but the target feature is continuous. In forecasting tasks, the target feature typically represents a future value estimated from historical observations at a defined prediction horizon. In time series regression tasks, the target feature may instead describe a continuous property or external response associated with the sequence. We treat both cases treated as regression because they require the prediction and can be addressed using the same modeling approaches \cite{hyndman_forecasting_2018-1,haben_time_2023,mohammadi_foumani_deep_2024}.

\vspace{-0.5em}
\section{Methodology}
\label{sec:methodology}

\begin{figure}[htbp]
    \centering
    \makebox[\textwidth][c]{%
        \includegraphics[width=1.2\textwidth]{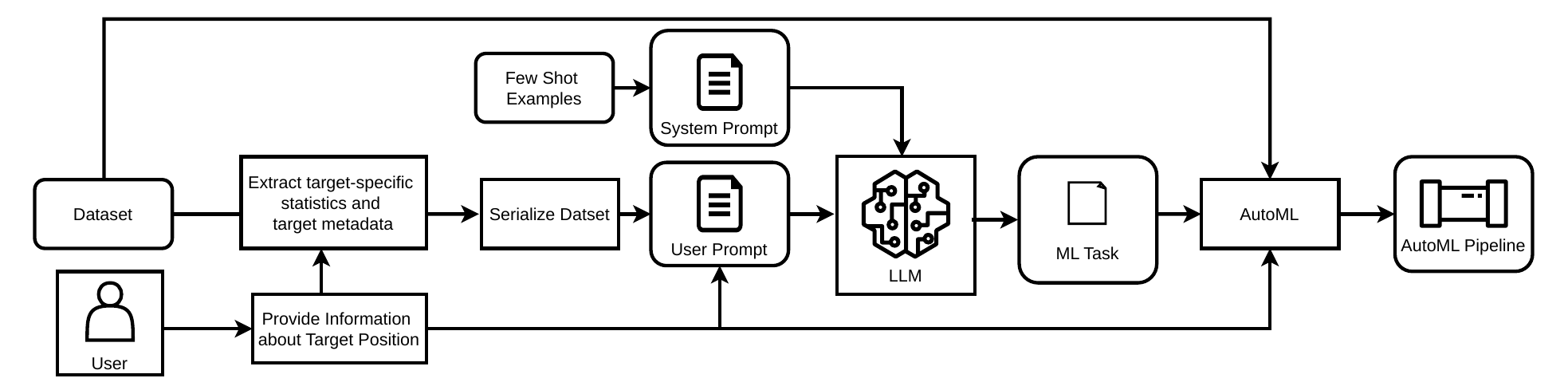}
    }
    \caption{Workflow of the proposed system. Given a dataset, target feature, and optional textual description, the system extracts target-specific statistics, generates a compact serialized representation, and combines the information into a prompt for predicting the data domain and downstream ML task via the LLM.}
    \label{fig:final}
\end{figure}

The proposed system identifies both the data domain and the downstream ML task from structured dataset information and optional semantic context. As shown in Figure~\ref{fig:final}, the workflow starts with two mandatory inputs: the dataset and an explicitly specified target feature, which defines the prediction objective and cannot easily be inferred automatically. When available, a textual dataset description is added as semantic context.

\vspace{-0.2em}
\paragraph{Target-specific statistics}
Before prompt construction, task-relevant statistics are extracted from the target variable, including value counts, unique values, mean, standard deviation, minimum, and maximum. For text-like targets, word, stopword, and character counts are additionally extracted with their standard deviations. These statistics provide compact information about the prediction objective.


\paragraph{Dataset serialization}
The dataset itself is then serialized into a structured DataFrame representation that can be processed by the LLM, as shown in Appendix~\ref{app:prompt-templates}. The representation preserves the row- and column-based dataset structure, including original feature names, which can provide semantic cues about the variables. The target feature name is always retained and explicitly marked in the prompt because it defines the prediction objective. Several serialization strategies were explored, and the final setup employs this compact matrix-like representation to preserve the original structure while remaining suitable for prompt-based input \cite{aghajanyan_htlm_2021,sui_table_2024, eisenschlos_mate_2021, iida_tabbie_2021,liu_tapex_2021, deng_turl_2022, herzig_tapas_2020, gong_tablegpt_2020,nassar_tableformer_2022,wang_tuta_2021}.

\paragraph{Group-wise row sampling}
To reduce computational cost and avoid exceeding the model context window, each dataset is represented by a subset of selected samples and features \cite{dong_survey_2024, jaitly_towards_2023}. We use group-wise random sample selection, where contiguous sample groups are selected without replacement and then reordered by their original positions. This preserves short-range patterns relevant to structured data, such as time series or grouped records, while keeping the input tractable. The target feature is always retained.


\paragraph{Prompting structure}

Prompt design is central to the proposed system because it strongly affects LLM performance. As shown in Appendix~\ref{app:prompt-templates}, the prompting framework combines a system prompt, which defines the task and output format, with a user prompt containing the serialized dataset, target feature information, extracted target-specific statistics, and optional dataset description. In few-shot settings, task-specific labeled examples from the training data are added to provide representative guidance for each downstream task. We evaluate both zero-shot and few-shot prompting. Zero-shot prompting uses only the system and dataset-specific user prompt, whereas few-shot prompting additionally includes labeled examples.

\paragraph{Data split and hyperparameter selection}
As shown in Table~\ref{tab:hyperparameter_space}, the experimental hyperparameter space includes the underlying LLM, reasoning mode, prompting strategy, and dataset-specific information configuration. GPT-5.3 serves as the cloud-based state-of-the-art model, while Qwen provides locally deployable alternatives with demonstrated strength in structured-data and reasoning tasks \cite{lin_benchmarking_2025}. For GPT-5.3, the reasoning effort was set to high in all experimental configurations. We include both recent Qwen3 \cite{wang_adaptable_2024} models and the established Qwen2.5-14B. The larger 14B models target high-memory industrial GPUs, such as the NVIDIA H100, whereas the compact 4B FP8 variants are intended for deployment on consumer-grade hardware, such as the NVIDIA RTX 5090. We compare zero-shot and few-shot prompting and vary the provided dataset information, ranging from the serialized dataset alone to additional target-specific statistics and optional dataset descriptions. The target feature is specified in all configurations, and hyperparameter optimization is conducted independently for each experiment. 
For configuration selection, the dataset collection is split into stratified training, validation, and test partitions of 20\%, 20\%, and 60\%. Since the foundation models remain fixed, this split separates method development, configuration selection, and final evaluation rather than model-parameter training. Because we do not tune model parameters, the training split can be substantially smaller than in traditional machine learning settings. The training split is used for manual method development, including prompt engineering for the system and user prompts, adaptation of the dataset serialization strategy, selection of few-shot examples, and identification of the most informative target-specific statistics. The validation split is then used for model selection and dataset-configuration selection, including the choice of model, in-context learning strategy, reasoning mode, and input information setting. Finally, the selected configurations are evaluated once on the held-out test split. The comparatively large test partition provides a broad and statistically more stable basis for the final evaluation across heterogeneous datasets. We do not use conventional cross-validation or nested cross-validation because the tuning process on the training split involves partially manual design decisions, particularly prompt engineering, dataset serialization, and representation choices. Repeating these development steps independently for multiple cross-validation folds would require substantial manual effort and would make it difficult to ensure that each fold is optimized consistently and without introducing experimenter bias. Instead, once the methodology has been finalized using the training split, we assess the statistical robustness of the reported results by applying bootstrap resampling independently to the validation and test splits. Specifically, we generate 1,000 bootstrap samples by sampling with replacement from each evaluation split and recompute the performance metrics for every sample. The reported results correspond to the mean and standard deviation across these bootstrap samples, providing statistically robust performance estimates while preserving the fixed train/validation/test protocol. 

\begin{table}[htbp]
    \centering
    \caption{Hyperparameter space evaluated in the LLM experiments from January to March 2026, including model choice, reasoning mode, prompting strategy, and dataset information configuration.}
    \label{tab:hyperparameter_space}
    \vspace{-0.2em}
    \setlength{\tabcolsep}{6pt}
    \renewcommand{\arraystretch}{1.1}
    \begin{tabularx}{\textwidth}{p{0.25\textwidth} X}
        \toprule
        \textbf{Hyperparameter} & \textbf{Evaluated values} \\[0.2em]
        Model &
        GPT-5.3; Qwen3-14B; Qwen2.5-14B-Instruct; Qwen3-4B-Instruct-2507-FP8; Qwen3-4B-Thinking-2507-FP8 \\
        
        Reasoning &
        \cmark; \xmark \\
        
        Prompting strategy &
        Zero-shot; Few-shot \\
        
        Dataset information configuration &
        Target feature name; target feature name + dataset description; target feature name + dataset description + target-specific statistics \\
        \bottomrule
    \end{tabularx}
\end{table}

\vspace{-0.8em}
\section{Experimental Results}
\label{sec:result}

\paragraph{Experimental setup}

\begin{figure}[!htbp]
    \centering
    \setlength{\abovecaptionskip}{2pt} 
    \setlength{\belowcaptionskip}{0pt}
    \captionsetup[subfigure]{skip=2pt}

    \makebox[\textwidth][c]{%
    \begin{minipage}{1.1\textwidth}
        \centering

        \begin{subfigure}[t]{.32\linewidth}
            \centering
            \includegraphics[
                width=\linewidth,
                trim={0.4cm 0.15cm 0.2cm 0.15cm},
                clip
            ]{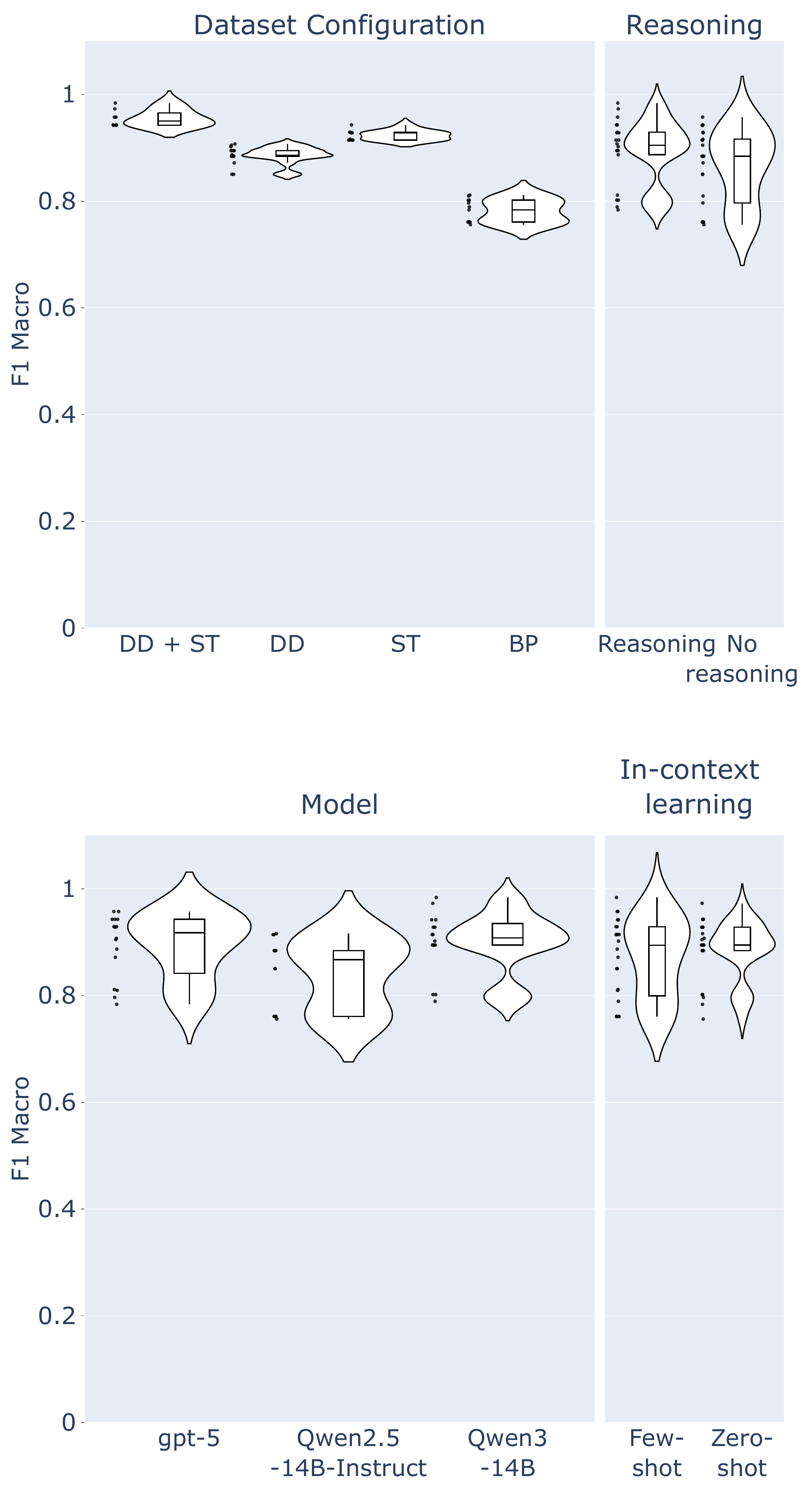}
            \caption{Experiment 1}
            \label{fig:violin_exp1}
        \end{subfigure}\hfill
        \begin{subfigure}[t]{.32\linewidth}
            \centering
            \includegraphics[
                width=\linewidth,
                trim={0.4cm 0.15cm 0.2cm 0.15cm},
                clip
            ]{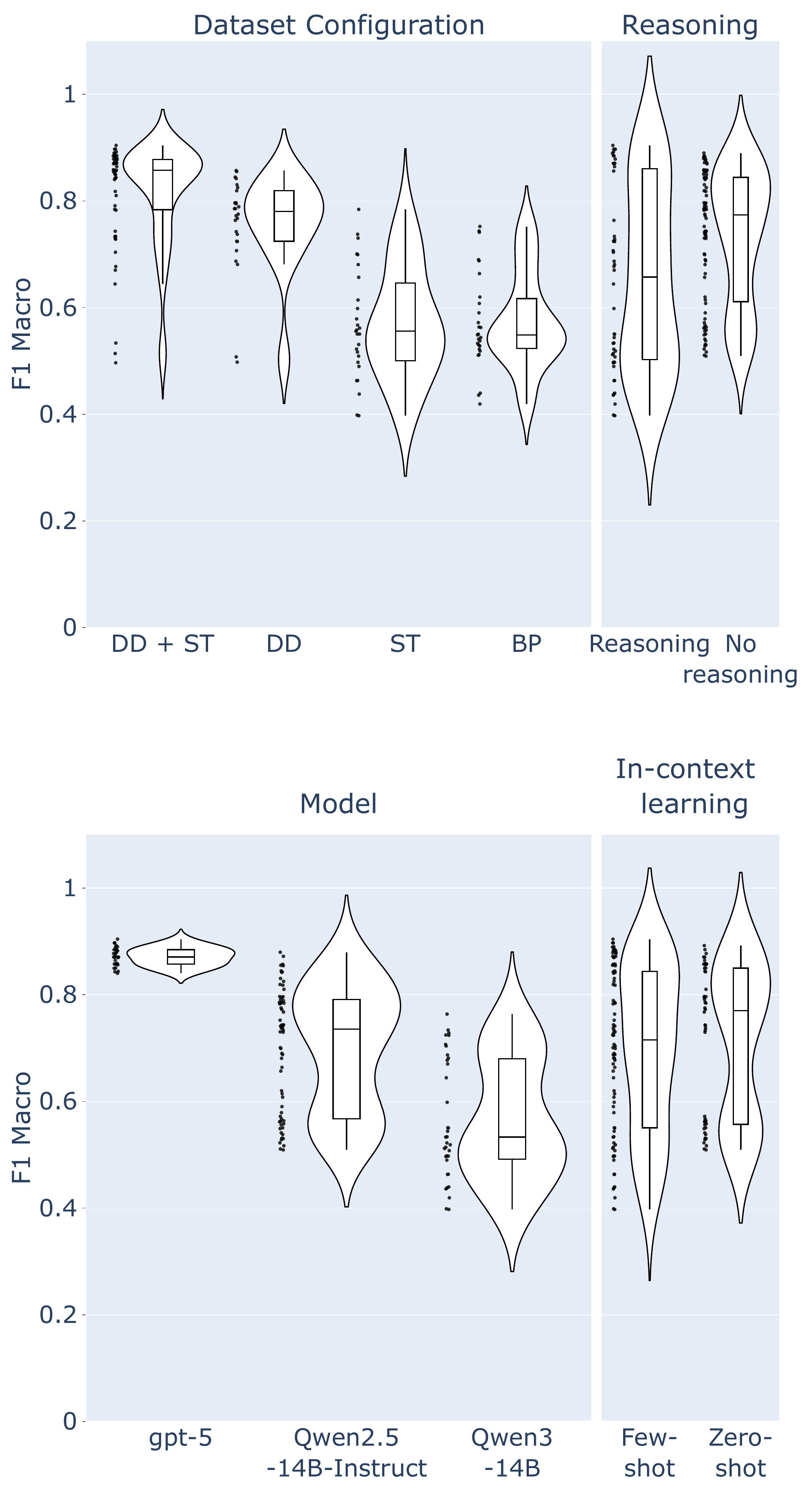}
            \caption{Experiment 2}
            \label{fig:violin_exp2}
        \end{subfigure}\hfill
        \begin{subfigure}[t]{.32\linewidth}
            \centering
            \includegraphics[
                width=\linewidth,
                trim={0.4cm 0.15cm 0.2cm 0.15cm},
                clip
            ]{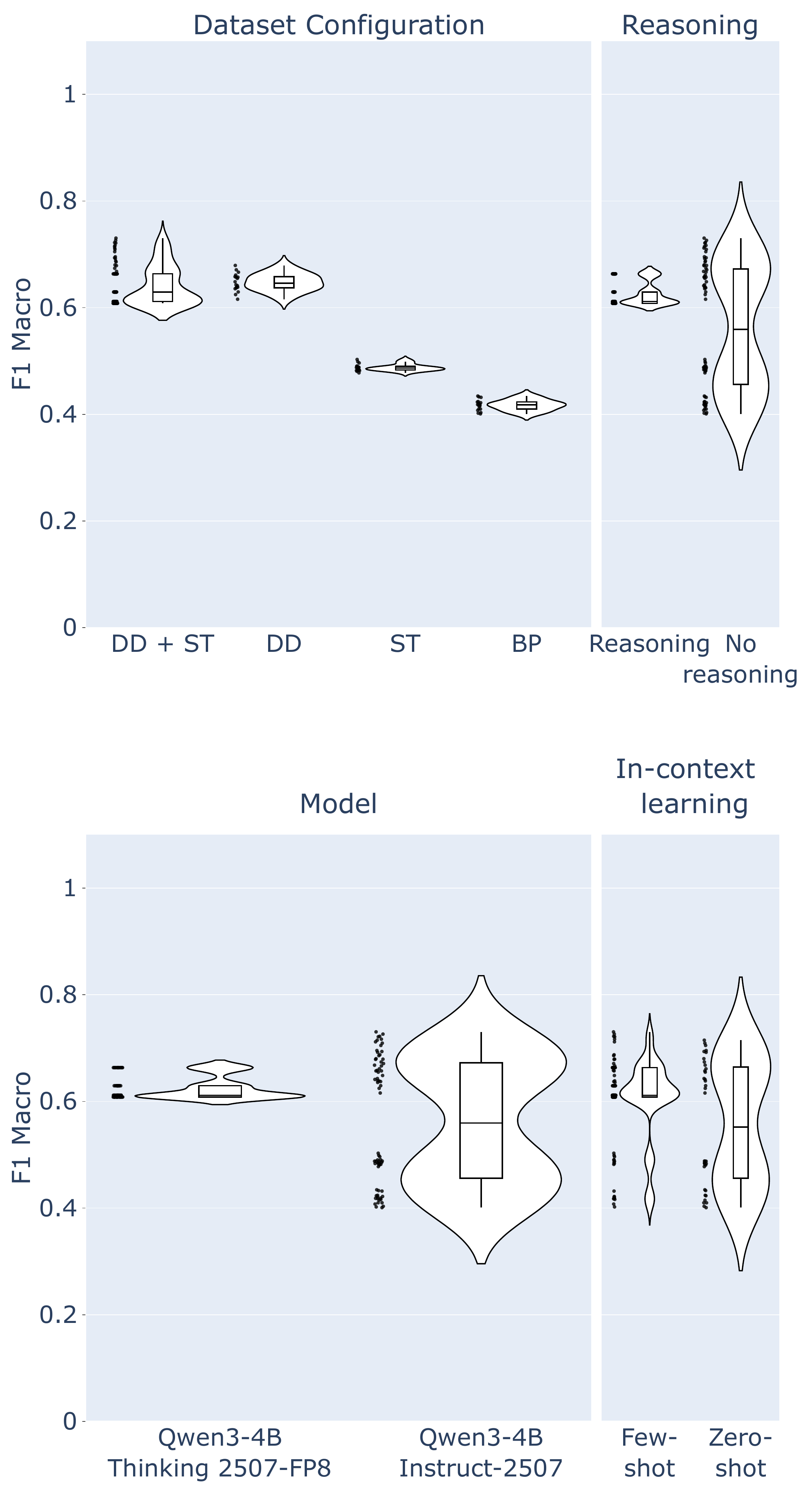}
            \caption{Experiment 3}
            \label{fig:violin_exp3}
        \end{subfigure}

    \end{minipage}%
    }

    \vspace{0.4em}

    \caption{F1-macro distributions across validation runs for Experiments 1–3. 
    Each violin aggregates all runs for a given hyperparameter across the remaining ones. 
    BP denotes the base prompt, DD dataset descriptions, ST target-specific statistics, and DD+ST their combination.}
    \label{fig:combined_violinplots_all}
\end{figure}

The evaluation follows the three research questions from Section~\ref{sec:introduction}: Experiment~1 compares the proposed approach against three established AutoML frameworks, namely AutoGluon, H2O, and NaiveAutoML, on tabular datasets. Experiment~2 evaluates cross-domain identification on tabular and time series datasets and Experiment~3 assesses smaller locally deployable models. We restrict the comparison with AutoML frameworks to tabular datasets because AutoGluon, H2O, NaiveAutoML, and most other AutoML frameworks only support task type identification for tabular data
Validation results are summarized in Figure~\ref{fig:combined_violinplots_all}, selected configurations are evaluated once on the held-out test split. 
The results report both DD+ST and ST-only settings, where DD denotes the textual dataset description and ST denotes target-specific statistics extracted from the target feature. This comparison contrasts the full information setting with the use of dataset-derived target-specific statistics alone.

\vspace{-0.4em}
\paragraph{Experiment 1}
As shown in Figure~\ref{fig:combined_violinplots_all}, the best validation performance in the tabular-only setting is achieved with DD+ST, followed closely by ST alone. 
GPT-5 achieves the strongest validation performance, Qwen3-14B remains competitive, and Qwen2.5-14B performs weakest, while reasoning mode and prompting strategy have only minor effects.

    

\begin{table}[!htbp]
    \centering
    \caption{Test-set results of the best-performing models in Experiment~1 for tabular ML task identification compared to the AutoGluon baseline. Performance metrics for the LLM-based methods are reported as the mean and standard deviation over 1{,}000 bootstrap resamples. DD denotes textual dataset descriptions and ST denotes target-specific statistics.}
    \vspace{-0.6em}
    \label{tab:exp1_results}
    \setlength{\tabcolsep}{5pt}
    \renewcommand{\arraystretch}{1.0}

    \begin{tabular}{lllcccc}
        \toprule
        & & & \multicolumn{2}{c}{Few-shot} & \multicolumn{2}{c}{Zero-shot} \\
        Model & Reasoning & Info. & F1 Macro & Balanced Acc. & F1 Macro & Balanced Acc. \\
        \midrule

        \textbf{AutoGluon} & -- & -- &
        --  &
        -- &
        \textbf{0.93 $\pm$ 0.01} &
        0.93 $\pm$ 0.01 \\

        NaiveAutoML & -- & -- &
        -- &
        -- &
        0.80 $\pm$ 0.01 &
        0.81 $\pm$ 0.01 \\

        H2O & -- & -- &
        -- &
        -- &
        0.59 $\pm$ 0.01 &
        0.58 $\pm$ 0.01 \\

        \midrule

        Qwen3-14B & \xmark & DD+ST &
        0.98 $\pm$ 0.01 &
        0.98 $\pm$ 0.01 &
        \textbf{0.98 $\pm$ 0.01} &
        \textbf{0.98 $\pm$ 0.00} \\

        Qwen3-14B & \cmark & DD+ST &
        0.98 $\pm$ 0.00 &
        0.98 $\pm$ 0.00 &
        0.98 $\pm$ 0.01 &
        0.98 $\pm$ 0.01 \\

        GPT-5.3 & \cmark & DD+ST &
        0.97 $\pm$ 0.00 &
        0.97 $\pm$ 0.00 &
        0.97 $\pm$ 0.01 &
        0.97 $\pm$ 0.01 \\

        GPT-5.3 & \xmark & DD+ST &
        0.97 $\pm$ 0.01 &
        0.97 $\pm$ 0.01 &
        0.97 $\pm$ 0.02 &
        0.97 $\pm$ 0.01 \\

        \midrule

        Qwen3-14B & \xmark & ST &
        0.96 $\pm$ 0.01 &
        0.96 $\pm$ 0.00 &
        0.96 $\pm$ 0.01 &
        0.96 $\pm$ 0.01 \\

        Qwen3-14B & \cmark & ST &
        \textbf{0.96 $\pm$ 0.01} &
        \textbf{0.96 $\pm$ 0.01} &
        0.95 $\pm$ 0.02 &
        0.96 $\pm$ 0.01 \\

        GPT-5.3 & \cmark & ST &
        0.97 $\pm$ 0.00 &
        0.97 $\pm$ 0.00 &
        0.96 $\pm$ 0.00 &
        0.97 $\pm$ 0.00 \\

        GPT-5.3 & \xmark & ST &
        0.96 $\pm$ 0.01 &
        0.96 $\pm$ 0.01 &
        0.96 $\pm$ 0.01 &
        0.96 $\pm$ 0.01 \\

        \bottomrule
    \end{tabular}
\end{table}

\begin{figure}[htbp]
    \centering

    \begin{subfigure}[t]{0.49\textwidth}
        \centering
        \includegraphics[width=\linewidth,
        trim = 0cm 0.85cm 0cm 0.65cm,
        clip]
        {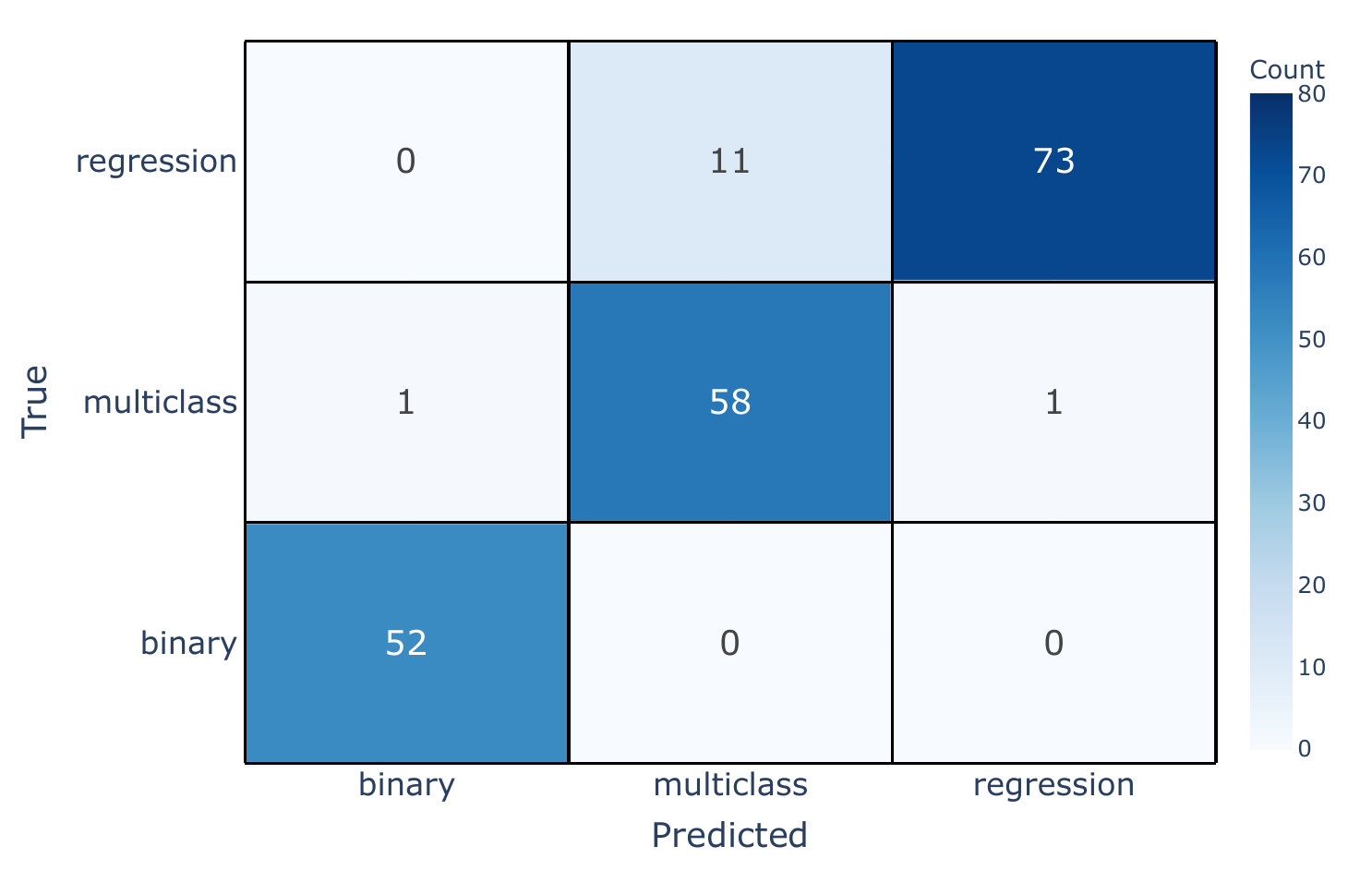}
        \caption{AutoGluon}
        \label{fig:exp1_cm_autogluon}
    \end{subfigure}
    \hfill
    \begin{subfigure}[t]{0.49\textwidth}
        \centering
        \includegraphics[width=\linewidth,
        trim = 0cm 0.85cm 0cm 0.65cm,
        clip]
        {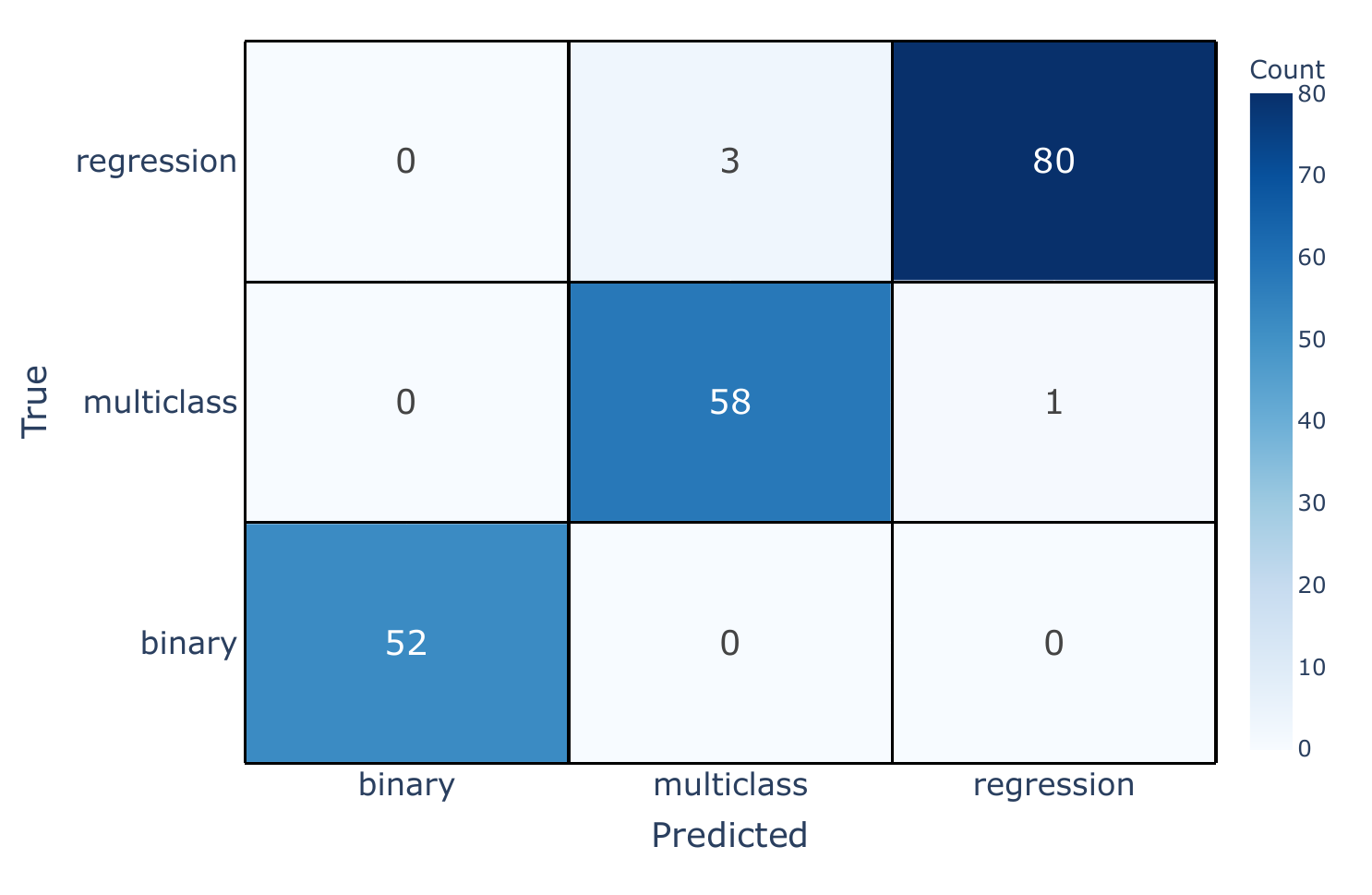}
        \caption{Qwen3-14B}
        \label{fig:exp1_cm_qwen3_14b}
    \end{subfigure}

    \vspace{-0.8em}
    \caption{Test-set confusion matrices for Experiment~1. \subref{fig:exp1_cm_autogluon}: AutoGluon baseline. \subref{fig:exp1_cm_qwen3_14b}: selected Qwen3-14B configuration.}
    \label{fig:exp1_confusion_matrices}
\end{figure}

Based on these results, the primary test evaluation uses DD+ST, with ST-only results reported for comparison. Table~\ref{tab:exp1_results} shows that all LLM-based configurations outperform AutoGluon. While AutoGluon reaches 0.93 F1 macro, DD+ST configurations reach 0.97--0.98, with Qwen3-14B achieving 0.98 across all prompting and reasoning settings. ST-only configurations also outperform AutoGluon, reaching 0.95--0.97 F1 macro, indicating that target-specific statistics already provide strong signals for tabular task identification. Figure~\ref{fig:exp1_confusion_matrices} further shows that AutoGluon has a tendency to confuse regression with multiclass classification, whereas the selected LLM configuration produces only rare misclassifications.
without caption

\paragraph{Experiment 2}
Figure~\ref{fig:combined_violinplots_all} indicates that validation performance across tabular and time series datasets is highest when semantic dataset information is included. DD+ST performs best, followed closely by DD alone, which also shows lower variance. In contrast, ST alone and the base prompt lead to lower and more variable performance, suggesting that target-specific statistics are useful for some datasets but less reliable across the heterogeneous benchmark. This indicates that cross-domain task identification depends more strongly on semantic context than the tabular-only setting, since the model must infer both the downstream task and the data domain. GPT-5.3 achieves the strongest validation performance, followed by Qwen2.5-14B and Qwen3-14B. Zero-shot prompting and reasoning show slightly more stable overall results, but high-performing configurations occur across several settings, indicating that model choice and dataset information have stronger effects. The primary test evaluation therefore uses DD+ST, with ST-only results reported for comparison. As shown in Table~\ref{tab:model_results_exp2}, GPT-5.3 with reasoning achieves the best DD+ST result, reaching 0.90 F1 macro in the few-shot setting. GPT-5.3 without reasoning reaches 0.86 and 0.84 F1 macro in the few-shot and zero-shot settings, respectively, indicating that reasoning is beneficial for the strongest cloud-based model. Among the local models, Qwen2.5-14B performs best with 0.84 F1 macro in the few-shot setting, while Qwen3-14B reaches 0.76 and 0.77 in the few-shot and zero-shot settings.

\begin{table}[!htbp]
    \centering
    \caption{Test-set results of the best-performing models in Experiment~2 for cross-domain ML task identification across tabular and time series datasets. Performance metrics are reported as the mean and standard deviation over 1{,}000 bootstrap resamples. DD denotes textual dataset descriptions and ST denotes target-specific statistics.}
    \label{tab:model_results_exp2}
    \vspace{-0.6em}
    \setlength{\tabcolsep}{5pt}
    \renewcommand{\arraystretch}{1.0}

    \begin{tabular}{lllcccc}
        \toprule
        & & & \multicolumn{2}{c}{Few-shot} & \multicolumn{2}{c}{Zero-shot} \\
        Model & Reasoning & Info. & F1 Macro & Balanced Acc. & F1 Macro & Balanced Acc. \\
        \midrule

        Qwen2.5-14B-Inst. & \xmark & DD+ST
        & 0.84 $\pm$ 0.04
        & 0.85 $\pm$ 0.03
        & 0.77 $\pm$ 0.04
        & 0.77 $\pm$ 0.04 \\

        Qwen3-14B & \cmark & DD+ST
        & 0.76 $\pm$ 0.02
        & 0.76 $\pm$ 0.02
        & 0.77 $\pm$ 0.02
        & 0.77 $\pm$ 0.01 \\

        GPT-5.3 & \cmark & DD+ST
        & \textbf{0.90 $\pm$ 0.03}
        & \textbf{0.90 $\pm$ 0.03}
        & 0.85 $\pm$ 0.03
        & 0.86 $\pm$ 0.01 \\

        GPT-5.3 & \xmark & DD+ST
        & 0.86 $\pm$ 0.02
        & 0.86 $\pm$ 0.01
        & 0.84 $\pm$ 0.03
        & 0.84 $\pm$ 0.03 \\

        \midrule

        Qwen2.5-14B-Inst. & \xmark & ST
        & 0.74 $\pm$ 0.03
        & 0.74 $\pm$ 0.04
        & 0.52 $\pm$ 0.06
        & 0.59 $\pm$ 0.04 \\

        Qwen3-14B & \cmark & ST
        & 0.68 $\pm$ 0.04
        & 0.69 $\pm$ 0.04
        & 0.71 $\pm$ 0.05
        & 0.70 $\pm$ 0.04 \\

        GPT-5.3 & \cmark & ST
        & 0.86 $\pm$ 0.03
        & 0.86 $\pm$ 0.02
        & \textbf{0.87 $\pm$ 0.02}
        & \textbf{0.88 $\pm$ 0.03} \\

        GPT-5.3 & \xmark & ST
        & 0.83 $\pm$ 0.01
        & 0.84 $\pm$ 0.01
        & 0.72 $\pm$ 0.03
        & 0.73 $\pm$ 0.02 \\

        \bottomrule
    \end{tabular}
\end{table}

\begin{figure}[htbp]
    \centering

    \begin{subfigure}[t]{0.49\textwidth}
        \centering
        \includegraphics[width=\linewidth,
        trim = 0cm 0.25cm 0cm 0.65cm,
        clip]
        {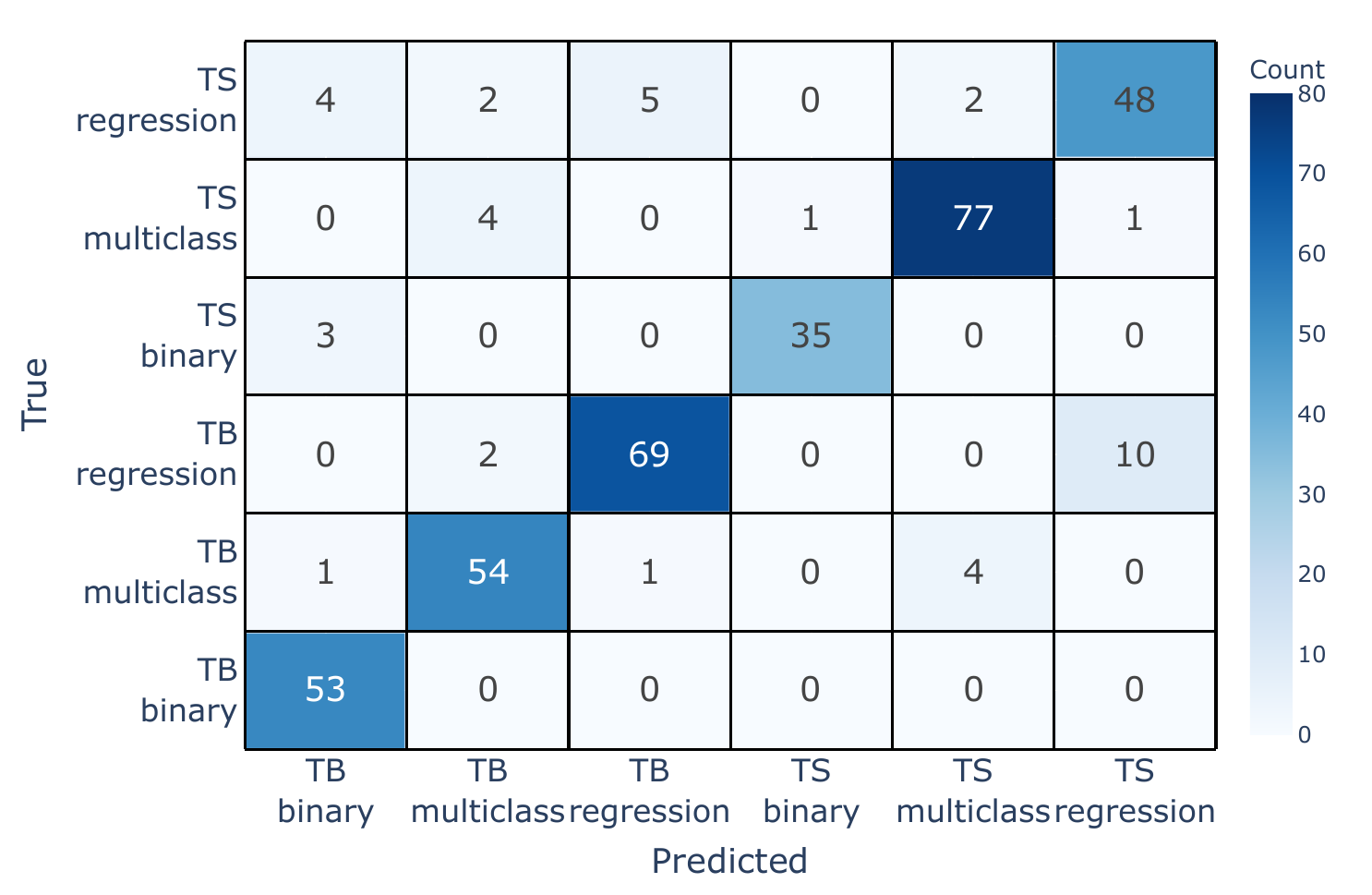}
        \caption{Exp. 2: GPT-5}
        \label{fig:cm_exp2_gpt5}
    \end{subfigure}
    \hfill
    \begin{subfigure}[t]{0.49\textwidth}
        \centering
        \includegraphics[width=\linewidth,
        trim = 0cm 0.25cm 0cm 0.65cm,
        clip]
        {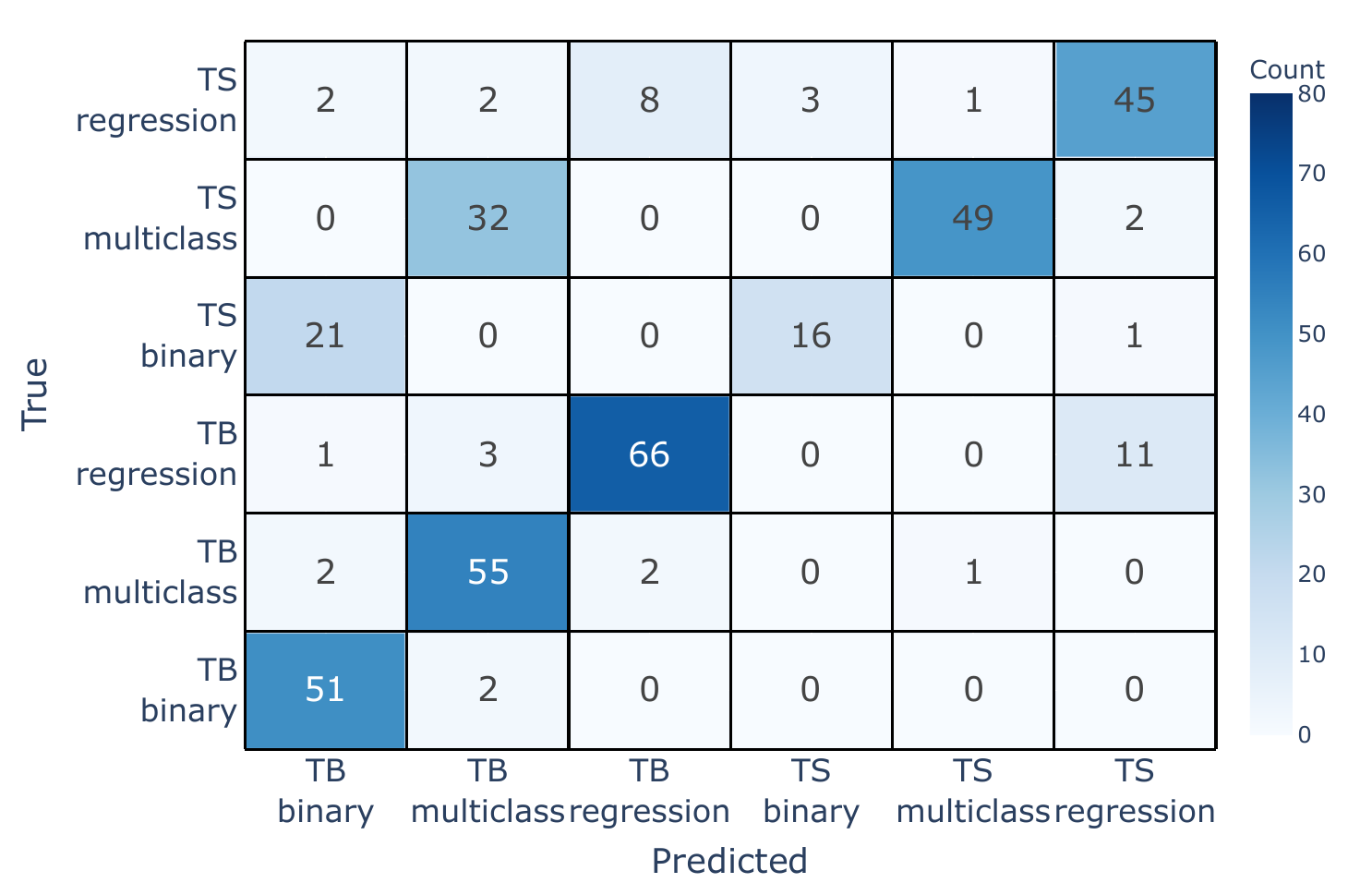}
        \caption{Exp. 3: Qwen-4B}
        \label{fig:cm_exp3_qwen3_4b}
    \end{subfigure}

    \caption{Confusion matrices of the best-performing models in the cross-domain setting. \subref{fig:cm_exp2_gpt5}: best state-of-the-art model in Experiment~2. \subref{fig:cm_exp3_qwen3_4b}: best locally deployable model in Experiment~3. TS denotes time series, and TB denotes tabular.}
    \label{fig:confusion_matrices_selected}
\end{figure}

ST-only results show that target-specific statistics can already support task identification, but less robustly across models. GPT-5.3 with reasoning reaches 0.86 and 0.87 F1 macro in the few-shot and zero-shot settings, respectively, whereas Qwen2.5-14B drops to 0.74 and 0.52. Thus, larger models exploit compact target-specific statistics more effectively, while smaller models benefit more from additional semantic context. Overall, Experiment~2 shows that cross-domain task identification is harder than the tabular-only setting, mainly because the model must also recognize the underlying data structure. This is also evident from the confusion matrix in Figure~\ref{fig:confusion_matrices_selected}, where the most frequent confusions occur between the same downstream task across the two domains. Nevertheless, LLM-based methods generalize across heterogeneous modalities, with DD+ST remaining the most robust configuration.

\paragraph{Experiment 3}
As shown in Figure~\ref{fig:combined_violinplots_all}, the best validation performance for smaller local models is achieved with DD+ST. ST provides useful information, but is less reliable as the only source of task type identification evidence. Since Experiment~3 includes only two local models, one with reasoning and one without, model choice and reasoning mode cannot be separated. Few-shot prompting performs slightly better than zero-shot prompting, but both are retained for test evaluation.

Table~\ref{tab:model_results_exp3} shows that smaller locally deployable models achieve meaningful cross-domain performance, although below the larger models from Experiment~2. With DD+ST, Qwen3-4B-Instruct performs best, reaching 0.75 F1 macro and 0.74 balanced accuracy in the few-shot setting, and 0.73 for both metrics in the zero-shot setting. Qwen3-4B-Thinking performs substantially worse, reaching 0.62 F1 macro in few-shot and 0.45 in zero-shot, indicating that the reasoning-oriented variant is less robust at this model scale. 

\begin{table}[!htbp]
\centering
\caption{Test-set results of the best-performing locally deployable models in Experiment~3 for cross-domain task identification under resource-constrained settings. Performance metrics are reported as the mean and standard deviation over 1{,}000 bootstrap resamples. DD denotes textual dataset descriptions and ST denotes target-specific statistics.}
\vspace{-0.6em}
\label{tab:model_results_exp3}
\setlength{\tabcolsep}{4pt}
\renewcommand{\arraystretch}{1.05}

\begin{tabular}{llccccc}
\toprule
& & & \multicolumn{2}{c}{Few-shot} & \multicolumn{2}{c}{Zero-shot} \\
Model & Reasoning & Info. & \shortstack{F1 Macro} & \shortstack{Balanced Acc.} & \shortstack{F1 Macro} & \shortstack{Balanced Acc.} \\
\midrule
Qwen3-4B-Instruct & \xmark & DD+ST & \textbf{0.75 $\pm$ 0.02} & \textbf{0.74 $\pm$ 0.02} & 0.73 $\pm$ 0.02 & 0.73 $\pm$ 0.02 \\
Qwen3-4B-Thinking & \cmark & DD+ST & 0.62 $\pm$ 0.03 & 0.64 $\pm$ 0.03 & 0.45 $\pm$ 0.04 & 0.45 $\pm$ 0.04\\
\midrule
Qwen3-4B-Instruct & \xmark & ST & \textbf{0.50 $\pm$ 0.01} & \textbf{0.58 $\pm$ 0.02} & 0.48 $\pm$ 0.03 & 0.56 $\pm$ 0.02 \\
Qwen3-4B-Thinking & \cmark & ST & 0.46 $\pm$ 0.02 & 0.55 $\pm$ 0.02 & 0.45 $\pm$ 0.04 & 0.48 $\pm$ 0.04 \\
\midrule
\end{tabular}
\end{table}

ST-only results lead to a clear performance drop for both local models. Qwen3-4B-Instruct-decreases from 0.75 to 0.50 F1 macro in few-shot and from 0.73 to 0.48 in zero-shot, while Qwen3-4B-Thinking remains near the lower performance range. This suggests that small local models require semantic context to resolve ambiguities between tabular and time series structures. Practical deployment is therefore feasible, but involves a clear trade-off between model size and task-identification accuracy. Detailed runtime and API-cost results are provided in Appendix~\ref{app:datasets}, together with token-usage statistics. 
The confusion matrix in Figure~\ref{fig:confusion_matrices_selected}
of the best locally deployable model in Experiment~3 shows the same error pattern as the best state-of-the-art model in Experiment~2, but with more frequent misclassifications. While it shows strong separation between regression and classification, most errors arise from confusion between the domains (tabular vs. time series). Hence, the local model often identifies the ML task correctly but struggles with domain identification.

\FloatBarrier

\section{Discussion}

\label{sec:discussion}
The results show that LLM-based approaches can effectively identify machine learning task types within the proposed taxonomy. In the tabular setting, they outperform the heuristic AutoML baseline, indicating that structural and semantic dataset information improves task inference beyond rule-based criteria. In the cross-domain setting, performance decreases once time series datasets are included, mainly because the model must also recognize the underlying data domain. This highlights domain identification as an important component for future AutoML systems.

The experiments further show that the input representation is crucial. 
The most stable results are obtained when the serialized dataset preserves the original row- and column-based structure, including feature names and an explicitly marked target feature, and is enriched with dataset descriptions and target-specific statistics. 
At the same time, target-specific statistics alone still achieve competitive results, especially for stronger state-of-the-art models. 
This indicates that task-relevant signals can be inferred directly from the dataset without textual descriptions. 
Nevertheless, in the cross-domain setting, the inclusion of dataset descriptions leads to some improvements in performance and reduces variance also for the stronger models. 
In contrast, smaller local models are less robust when only target-specific statistics are provided, suggesting that they depend more strongly on additional semantic context. Dataset descriptions are therefore not essential for accurate task identification but they improve the robustness, particularly in the cross-domain scenario and for the smaller-models. Overall, the results demonstrate that accurate task identification is feasible with minimal user-provided information, namely the target feature and, implicitly during the dataset creation phase, the feature names.

Several limitations remain. First, the benchmark mainly consists of curated public datasets and may not fully reflect the noise, inconsistency, and complexity of industrial data.Second, potential data contamination cannot be ruled out \cite{jiang_investigating_2024,balloccu_leak_2024}. This is a general limitation when evaluating pre-trained LLMs on public datasets, particularly for closed-source models with undisclosed training corpora. Public datasets, metadata, or related variants may have been included during pretraining, which can lead to improved performance. We mitigate this risk by excluding datasets listed in the LM Contamination Index\footnote{\url{https://hitz-zentroa.github.io/lm-contamination/}} and by defining the ML task-type labels independently of the original dataset repositories. Since these labels were manually assigned specifically for our benchmark and are not publicly available as part of the original datasets, they cannot have been directly memorized during pretraining, even if the corresponding dataset was included in the model's training corpus. However, contamination through dataset content or metadata remains possible, and future work should include dedicated contamination analyses, such as memorization tests or more comprehensive contamination indices. Third, the evaluation focuses on task type identification metrics such as F1 macro, accuracy, and balanced accuracy, but does not measure how misclassifications affect downstream pipeline stages. Future work should therefore quantify the impact of task type identification errors on preprocessing, model selection, training, evaluation, and overall pipeline performance. Automatic target-feature identification is another important extension, as it would remove the need for manual target specification.

\section{Conclusion}
\label{sec:conclusion}
This work introduced an LLM-based system and benchmark for machine learning task type identification. The results show that LLMs outperform heuristic AutoML task inference in the tabular setting and generalize to cross-domain identification across tabular and time series datasets. Smaller local models remain viable but require semantic context and show a clear accuracy--deployability trade-off. Overall, the findings support treating task identification as an explicit component of AutoML pipelines rather than as a fixed user-provided input.
\vspace{-0.2em}
\paragraph{Ethical considerations}
Automated task identification can help democratize AutoML by making ML workflows more accessible to non-expert users. However, incorrect predictions may propagate to later pipeline stages and lead to unsuitable preprocessing, model choices, or evaluation metrics, ultimately leading to biased or unreliable outcomes. The proposed system should therefore be used as a transparent, inspectable decision-support component under expert oversight.

\begin{acknowledgements}
This research was funded by the Honda Research Institute Europe, GmbH.
\end{acknowledgements}

\vspace{7em}

\bibliography{references}






\newpage
\section*{Submission Checklist}


\begin{enumerate}
\item For all authors\dots
  \begin{enumerate}
  \item Do the main claims made in the abstract and introduction accurately
    reflect the paper's contributions and scope?
    \answerYes{} The benchmark dataset is described in Section~\ref{sec:dataset}, 
    the LLM-based methodology is described in Section~\ref{sec:result}, 
    and the empirical analysis supporting the main claims is reported in 
    Section~\ref{sec:result}. The scope and implications of the 
    results are further discussed in Section~\ref{sec:discussion}.

  \item Did you describe the limitations of your work?
    \answerYes{} The limitations are discussed in Section~\ref{sec:discussion}. 
    These include the use of curated public datasets, the possibility of 
    benchmark contamination in foundation models, the focus on task type 
    identification metrics rather than downstream pipeline effects, and the 
    current assumption that the target feature is provided by the user.
   \item Did you discuss any potential negative societal impacts of your work?
    \answerYes{} Potential negative societal impacts are discussed in the 
    ethical considerations paragraph in Section~\ref{sec:conclusion}. In 
    particular, we note that incorrect task predictions may propagate to later 
    pipeline stages and lead to unsuitable preprocessing, model choices, or 
    evaluation metrics.
  \item Did you read the ethics review guidelines and ensure that your paper
    conforms to them? (see \url{https://2022.automl.cc/ethics-accessibility/})
    \answerYes{} The work follows the ethics guidelines, as also stated in the 
    ethical considerations paragraph in Section~\ref{sec:conclusion}. The system 
    is positioned as a transparent decision-support component that should remain 
    inspectable and subject to expert oversight.
  \end{enumerate}
\item If you ran experiments\dots
  \begin{enumerate}
  \item Did you use the same evaluation protocol for all methods being compared (e.g.,
    same benchmarks, data (sub)sets, available resources, etc.)?
    \answerYes{} All experiments follow the same train/validation/test split 
    strategy described in Section~\ref{sec:methodology}. The experiments are 
    aligned with the research questions and use comparable evaluation metrics 
    and procedures to ensure consistent comparison across methods and settings.
  \item Did you specify all the necessary details of your evaluation (e.g., data splits,
    pre-processing, search spaces, hyperparameter tuning details and results, etc.)?
    \answerYes{} The dataset construction and benchmark structure are described 
    in Section~\ref{sec:dataset}. The data splits, preprocessing and serialization 
    procedure, hyperparameter space, and configuration-selection procedure are 
    specified in Section~\ref{sec:methodology}. The corresponding empirical 
    results are reported in Section~\ref{sec:result}.
  \item Did you repeat your experiments (e.g., across multiple random seeds or
    splits) to account for the impact of randomness in your methods or data?
    \answerNo{} We use fixed stratified train, validation, and test splits rather than repeated random splits. The training set is used for prompt adaptation, the validation set for configuration selection, and the final selected configurations are evaluated once on a held-out test set.
    Repeated evaluations were not easily feasible, as the optimization process on the training and, to some extent, validation set involved manual iterative refinement that could not easily be reproduced in an unbiased manner. Furthermore, since we only used pretrained models without additional training or fine-tuning, one major source of variability was not under our control.
  \item Did you report the uncertainty of your results (e.g., the standard error
    across random seeds or splits)?
    \answerNo{} We do not report standard errors across repeated random seeds or 
    splits, since the experiments are based on fixed dataset partitions.
    However, in Fig.~\ref{fig:combined_violinplots_all} we show the distributions of the results of our experiments on the validation set, including the results for each individual data point. 
  \item Did you report the statistical significance of your results?
    \answerNo{} We do not report formal statistical significance tests. Instead, 
    we report macro-averaged F1 scores, balanced accuracy, and confusion matrices 
    across the benchmark datasets.
  \item Did you use enough repetitions, datasets, and/or benchmarks to support
    your claims?
    \answerYes{} We introduce and evaluate on a benchmark of 625 public tabular 
    and time series datasets, as described in Section~\ref{sec:dataset}. This 
    provides broad coverage for the considered task type identification setting.
  \item Did you compare performance over time and describe how you selected the
    maximum runtime?
     \answerNo{} Runtime-based performance comparison is not the focus of this 
    work. The evaluation focuses on task type identification accuracy across 
    dataset domains and model configurations.
  \item Did you include the total amount of compute and the type of resources
    used (e.g., type of \textsc{gpu}s, internal cluster, or cloud provider)?
    \answerYes{} The evaluated model families and the intended deployment 
    settings, including cloud-based models and locally deployable GPU settings, 
    are described in Section~\ref{sec:methodology}.
  \item Did you run ablation studies to assess the impact of different
    components of your approach?
    \answerYes{} We investigate the influence of several hyperparameters as part of the configuration-selection procedure, including model choice, In-context
learning strategy, reasoning mode, and dataset-information configuration. In particular, we analyze the impact of different dataset-information settings, such as the inclusion or exclusion of textual dataset descriptions. Since we did not perform additional model training or fine-tuning, we consider this systematic evaluation of inference-time configurations to be the most appropriate form of ablation study for our setting.
  \end{enumerate}
\item With respect to the code used to obtain your results\dots
  \begin{enumerate}
\item Did you include the code, data, and instructions needed to reproduce the
    main experimental results, including all dependencies (e.g.,
    \texttt{requirements.txt} with explicit versions), random seeds, an instructive
    \texttt{README} with installation instructions, and execution commands
    (either in the supplemental material or as a \textsc{url})?
    \answerYes{} The project repository contains the code, dependency 
    specification, execution instructions, and documentation required to 
    reproduce the main experiments. The dataset is provided through the dataset 
    repository. Both repositories are linked in Section~\ref{sec:introduction}.
  \item Did you include a minimal example to replicate results on a small subset
    of the experiments or on toy data?
    \answerNo{} We do not include a separate toy-data example. However, the 
    repository provides detailed instructions for running the code on the 
    released benchmark data. The default configuration uses a locally deployable 
    Qwen3-4B-Instruct model, which can be executed on consumer-grade GPU 
    hardware such as an RTX 5090.
  \item Did you ensure sufficient code quality and documentation so that someone else
    can execute and understand your code?
    \answerYes{} The code includes comments and documentation, and the 
    \texttt{README} provides detailed instructions for installing the 
    dependencies, accessing the dataset, and executing the experiments.
  \item Did you include the raw results of running your experiments with the given
    code, data, and instructions?
    \answerNo{} We do not include all raw model outputs. However, the fixed 
    dataset splits are released with the benchmark, and the experiments are 
    reproducible from the provided code, data, and instructions. Some variation 
    may remain due to nondeterminism in LLM outputs.
  \item Did you include the code, additional data, and instructions needed to generate
    the figures and tables in your paper based on the raw results?
    \answerYes{} The repository includes comparison and plotting utilities for 
    recreating the reported tables and figures. These utilities can be used 
    directly from the \texttt{utils} folder or imported after installing the 
    package as described in the \texttt{README}.
  \end{enumerate}
\item If you used existing assets (e.g., code, data, models)\dots
  \begin{enumerate}
  \item Did you cite the creators of used assets?
     \answerYes{} The creators and sources of the used public datasets, model 
    families, and relevant software frameworks are cited in the paper where 
    applicable.
  \item Did you discuss whether and how consent was obtained from people whose
    data you're using/curating if the license requires it?
    \answerNA{} The benchmark is constructed from publicly available data sources. 
Where applicable, dataset licenses and citation requirements were reviewed and 
the corresponding dataset sources are cited in the paper. We do not collect data 
directly from human participants.
  \item Did you discuss whether the data you are using/curating contains
    personally identifiable information or offensive content?
      \answerNA{} The work relies on public benchmark datasets and focuses on 
    dataset-level task type identification. We do not introduce new 
    person-level annotations or collect additional personally identifiable 
    information.
  \end{enumerate}
\item If you created/released new assets (e.g., code, data, models)\dots
  \begin{enumerate}
    \item Did you mention the license of the new assets (e.g., as part of your
    code submission)?
     \answerYes{} The released code and benchmark assets include license 
information in the corresponding GitHub repository and Zenodo dataset release.
    \item Did you include the new assets either in the supplemental material or as
    a \textsc{url} (to, e.g., GitHub or Hugging Face)?
    \answerYes{} The released code and benchmark dataset are provided through 
    the project repository and the dataset repository, which are linked in 
    Section~\ref{sec:introduction}.
  \end{enumerate}
\item If you used crowdsourcing or conducted research with human subjects\dots
  \begin{enumerate}
  \item Did you include the full text of instructions given to participants and
    screenshots, if applicable?
    \answerNA{} This work does not use crowdsourcing and does not involve human 
    subject experiments.
  \item Did you describe any potential participant risks, with links to
    institutional review board (\textsc{irb}) approvals, if applicable?
    \answerNA{} This work does not involve human participants.
  \item Did you include the estimated hourly wage paid to participants and the
    total amount spent on participant compensation?
    \answerNA{} No participants were recruited or compensated.
  \end{enumerate}
\item If you included theoretical results\dots
  \begin{enumerate}
  \item Did you state the full set of assumptions of all theoretical results?
    \answerNA{} This work does not include theoretical results.
  \item Did you include complete proofs of all theoretical results?
    \answerNA{} This work does not include theoretical results.
  \end{enumerate}
\end{enumerate}

\newpage
\appendix

\begin{center}
    {\LARGE\bfseries Appendix}
\end{center}
\vspace{1em}

\section{Additional Dataset Details}
\label{app:datasets}

\subsection{Repository Links}
\label{app:repository-links}

\begin{table}[htbp]
    \centering
    \caption{Repository links for the benchmark dataset, implementation, and LM Contamination Index.}
    \label{tab:repository_links}
    \vspace{-0.2em}
    \setlength{\tabcolsep}{6pt}
    \renewcommand{\arraystretch}{1.1}
    \begin{tabularx}{\textwidth}{p{0.28\textwidth} X}
        \toprule
        \textbf{Resource} & \textbf{Link} \\
        \midrule
        Benchmark dataset 
        & \url{https://zenodo.org/records/21649607} \\
        Project repository 
        & \url{https://github.com/ptsialis/Machine-Learning-Task-Type-Identification} \\
        
        LM Contamination Index 
        & \url{https://hitz-zentroa.github.io/lm-contamination/} \\
        \bottomrule
    \end{tabularx}
\end{table}
\subsection{Dataset Metadata}

In the benchmark dataset the original files are stored in a taxonomy-based folder structure.
In addition to the original dataset files, the benchmark includes a tabular meta-information dataset. This table stores high-level information for each dataset, including the dataset name, data domain, downstream task, target variable, dataset description, multi-target indicator, and target data type. For time series datasets, the metadata additionally records whether the data are univariate or multivariate and whether they contain one or multiple time series. General dataset properties are also included, such as the number of instances, the number of features, the dataset format, the relative storage path, and the original download link. Figure~\ref{fig:dataset_metadata_example} shows a representative metadata entry.

\subsection{Dataset Types}

The benchmark contains tabular and time series datasets. Tabular datasets are represented as collections of independent samples with feature columns and an explicitly specified target variable. They are assigned to binary classification, multiclass classification, or regression depending on the target variable and prediction objective.

Time series datasets contain observations with an intrinsic sequential or temporal structure. They include both univariate and multivariate time series, as well as datasets consisting of either a single time series or multiple time series instances. Time series classification datasets use categorical targets, whereas time series forecasting and regression datasets use continuous targets. In this work, forecasting is treated as a specific form of regression, since both require the prediction of continuous values.

\subsection{Runtime and Cost Considerations}

In addition to predictive performance, we report the practical access characteristics of the evaluated models, since deployment feasibility depends not only on accuracy but also on whether a model is accessed through a commercial API or executed locally. Table \ref{tab:model_latency_cost_by_prompting} summarizes the fee type, deployment-related latency, and monetary cost for each evaluated model and in-context learning strategy. 
Since the monetary cost of API-based models is largely determined by the number of processed input and output tokens, the corresponding token-usage statistics, including the average token consumption for the zero-shot and few-shot settings, are reported in Appendix Table~\ref{tab:token_usage_statistics}.  AutoGluon achieves by far the lowest latency, with an average inference time of only 0.0004s, making it several orders of magnitude faster than all LLM-based approaches. However, this speed advantage must be interpreted in relation to predictive performance. The locally deployed LLMs require longer inference times, but several configurations remain practically efficient, with latencies of only a few seconds per prompt. In particular, the non-thinking Qwen models show that local LLM-based inference can still be feasible for offline dataset analysis. Therefore, waiting a few additional seconds per dataset can be worthwhile when LLM-based methods provide improved classification performance, stronger semantic interpretation of dataset metadata, and better robustness across heterogeneous datasets. Moreover, the capabilities of compact LLMs have improved substantially in recent months despite their considerably smaller model sizes and lower computational requirements. This trend suggests that inference latency is likely to continue decreasing while maintaining competitive predictive performance, further improving the practicality of local LLM-based solutions.

\begin{table}[htbp]
    \centering
    \caption{Latency and cost by model, baseline, and in-context learning strategy. Latency denotes the average per-prompt inference time for locally deployed models and baselines. GPT-5.3 was evaluated using API batch processing to reduce costs, preventing comparable per-prompt latency measurement. Reported costs are the API charges incurred for each experimental setting between December 2025 and March 2026. GPT-5.3 is a commercial API model, whereas Qwen and AutoGluon incurred no API costs.}

    \label{tab:model_latency_cost_by_prompting}
    \vspace{-0.2em}
    \small
    \setlength{\tabcolsep}{4pt}
    \renewcommand{\arraystretch}{1.15}
    \begin{tabularx}{\textwidth}{p{0.13\textwidth} p{0.25\textwidth} p{0.16\textwidth} p{0.17\textwidth} X}
        \toprule
        \textbf{Fee type} 
        & \textbf{Model} 
        & \textbf{\shortstack{In-context\\learning}} 
        & \textbf{\shortstack{Avg. latency\\per prompt}} 
        & \textbf{Cost per prompt / total cost} \\
        \midrule

        \multirow{2}{*}{Paid API}
        & \multirow{2}{0.25\textwidth}{GPT-5.3}
        & Zero-shot 
        & -- 
        & \$0.003 / \$1.13 \\

        & 
        & Few-shot 
        & -- 
        & \$0.009 / \$3.38 \\
        \midrule

        Free
        & AutoGluon
        & -- 
        & $0.0004\,\mathrm{s}$
        & - \\
        
        Free
        & NaiveAutoML
        & -- 
        & $0.01\,\mathrm{s}$
        & - \\
        
        Free
        & H2O
        & -- 
        & $0.009\,\mathrm{s}$
        & - \\
        \midrule
        
        \multirow{2}{*}{Free}
        & \multirow{2}{0.25\textwidth}{Qwen3-14B}
        & Zero-shot 
        & $6.03\,\mathrm{s}$ 
        & - \\
        
        & 
        & Few-shot 
        & $17.02\,\mathrm{s}$
        & -\\

        \multirow{2}{*}{Free}
        & \multirow{2}{0.25\textwidth}{Qwen2.5-14B-Instruct}
        & Zero-shot 
        & $1.02\,\mathrm{s}$
        & - \\
        
        & 
        & Few-shot 
        & $7.04\,\mathrm{s}$
        & - \\
        
        \multirow{2}{*}{Free}
        & \multirow{2}{0.25\textwidth}{Qwen3-4B-Instruct-2507-FP8}
        & Zero-shot 
        & $8.39\,\mathrm{s}$
        & - \\
        
        & 
        & Few-shot 
        & $8.45\,\mathrm{s}$
        & - \\

        \multirow{2}{*}{Free}
        & \multirow{2}{0.25\textwidth}{Qwen3-4B-Thinking-2507-FP8}
        & Zero-shot 
        & $83.09\,\mathrm{s}$
        & - \\
        
        & 
        & Few-shot 
        & $417.13\,\mathrm{s}$
        & - \\
        \bottomrule
    \end{tabularx}
\end{table}

\subsection{Token Statistics}

The token usage statistics in Table~\ref{tab:token_usage_statistics} show the prompt lengths for the zero-shot and few-shot settings across all datasets. In the zero-shot setting, prompts contain 2,502 tokens on average, with a minimum of 537 and a maximum of 15,585 tokens. In the few-shot setting, the average prompt length increases to 7,400 tokens due to the additional in-context examples included in the system prompt. The user prompt remains unchanged between both settings, with an average length of 2,252 tokens.

\begin{table}[htbp]
    \centering
    \caption{Token usage statistics per prompt across all datasets.}
    \label{tab:token_usage_statistics}
    \vspace{-0.2em}
    \setlength{\tabcolsep}{4pt}
    \renewcommand{\arraystretch}{1.1}
    \begin{tabular}{lrrrrrrrrr}
        \toprule
        \textbf{Method} 
        & \multicolumn{3}{c}{\textbf{Total Tokens}} 
        & \multicolumn{3}{c}{\textbf{System Prompt Tokens}} 
        & \multicolumn{3}{c}{\textbf{User Prompt Tokens}} \\
        \cmidrule(lr){2-4}
        \cmidrule(lr){5-7}
        \cmidrule(lr){8-10}
        & \textbf{Min.} & \textbf{Max.} & \textbf{Avg.}
        & \textbf{Min.} & \textbf{Max.} & \textbf{Avg.}
        & \textbf{Min.} & \textbf{Max.} & \textbf{Avg.} \\
        \midrule
        Zero-shot 
        & 537 & 15585 & 2502
        & 238 & 238 & 238
        & 287 & 15335 & 2252 \\
        
        Few-shot 
        & 5435 & 20483 & 7400
        & 5136 & 5136 & 5136
        & 287 & 15335 & 2252 \\
        \bottomrule
    \end{tabular}
\end{table}

\subsection{Loading Pipeline}

The collected datasets were provided in heterogeneous file formats and storage structures, including CSV, XLS, ARFF, Arrow, and JSON files. This heterogeneity was especially pronounced for time series datasets, where data may be stored as single long sequences, collections of separate sequences, nested structures, or archive-specific formats.

To handle these differences, we implemented a loading pipeline that first identifies the dataset format and then applies a corresponding parser. The goal of the pipeline is not to convert all datasets into a single standardized schema, but to load each dataset into a processable representation while preserving its original structure as far as possible. This design choice reflects the intended benchmark setting: models should be evaluated on heterogeneous real-world dataset representations rather than on heavily homogenized inputs.

During loading, the pipeline also links each dataset to its metadata entry, including the target variable and task label. The explicitly specified target variable is retained in all downstream prompt representations, since the current benchmark focuses on identifying the data domain and downstream task given a known prediction target.

\definecolor{mygray}{RGB}{245,245,245}
\begin{figure}[H]
\centering
\begin{tcolorbox}[
    colback=mygray,
    colframe=black!30,
    boxrule=0.3pt,
    arc=1mm,
    left=1mm,
    right=1mm,
    top=0.5mm,
    bottom=0.5mm,
    title={Example Dataset from Collection},
    fonttitle=\bfseries\scriptsize,
    width=0.6\textwidth
]
\scriptsize
\setlength{\tabcolsep}{3pt}
\renewcommand{\arraystretch}{0.7}
\begin{tabularx}{\linewidth}{>{\bfseries}p{0.28\linewidth} X}
\toprule
Dataset Name & Liver Disorders \\
Domain Type & Tabular \\
Downstream Task & Regression \\
Multi-/Univariate & Tabular\_Multivariate \\
File Path & \texttt{Class\_Reg/liver-disorders/...} \\
Original Download Link & \url{https://openml.org/data/v1/download/8/liver-disorders.arff} \\
Target Variable & \texttt{drinks} \\
Multi Target & -- \\
Target Variable Data Type & \texttt{numpy.float64} \\
Features & 7 \\
Instances & 345 \\
Dataset Format & ARFF \\
Dataset Description & Dataset provided by BUPA Medical Research Ltd. with... \\
\bottomrule
\end{tabularx}
\end{tcolorbox}

\caption{\scriptsize High-level overview of a representative dataset in the collection.}
\label{fig:dataset_metadata_example}
\end{figure}

\section{Prompt Templates}
\label{app:prompt-templates}

\begin{promptbox}{System Prompt Used for Structured Data Classification}
You are a dataset classifier for structured data.
Your task is to identify:
1. The data domain (Tabular or Time Series)
2. The prediction task associated with the target variable.

### Task

#### Step 1: Identify the Data Domain
Classify the dataset as one of the following:
- 'Tabular' --- independent rows with no intrinsic temporal ordering.
- 'Time_Series' --- observations indexed or ordered by time, sequence, or temporal dependency.

Use dataset structure, column semantics, and context to determine the domain.

#### Step 2: Identify the Prediction Task

For Tabular data, classify the prediction task as one of:
- 'binary'
- 'multiclass'
- 'regression'

For Time Series data, classify the prediction task as one of:
- 'binary'
- 'multiclass'
- 'regression'

Rules for task identification:
- If the target variable is continuous -> 'regression'.
- If the target variable is categorical:
  - Exactly 2 unique values -> 'binary'.
  - More than 2 unique values -> 'multiclass'.
- Important: Integer-valued targets may be categorical or continuous.
  Decide based on semantic meaning and dataset context, not datatype alone.

### Input Format
You receive:
- A DFLoader-serialized excerpt of the dataset.
- A target specification in the following form:
  - "Target: column, <name>"

### Output Format
Respond only with:
('<Task Domain>', '<Sub Problem Task>')

Where:
- <Task Domain> is exactly one of:
  - 'Tabular'
  - 'Time_Series'
- <Sub Problem Task> is exactly one of:
  - 'binary'
  - 'multiclass'
  - 'regression'

Do not include explanations.
Do not include additional text.
\end{promptbox}

\begin{promptbox}{Example User Prompt}
Dataset description:
Author: BUPA Medical Research Ltd.; Donor: Richard S. Forsyth
Source: UCI Liver Disorders dataset, 5/15/1990

BUPA liver disorders:
The first five variables are blood tests that may indicate liver disorders
related to excessive alcohol consumption. Each row represents one male
individual.

Important note:
The seventh field, selector, is not a dependent variable. It was created by
BUPA researchers as a train/test selector and is not suitable as a
classification target. In this prompt, the sixth field, drinks, is used as
the target variable.

Attribute information:
1. mcv: mean corpuscular volume
2. alkphos: alkaline phosphotase
3. sgpt: alanine aminotransferase
4. sgot: aspartate aminotransferase
5. gammagt: gamma-glutamyl transpeptidase
6. drinks: half-pint equivalents of alcoholic beverages drunk per day
7. selector: train/test split field

Dataset:
pd.DataFrame({
    'mcv': [98, 88, 88, 92, 90, 89, 82, 90, 86, 96, 90, 87, 96, 91, 95, 94, 87, 98, 94, 83, 88, 82, 85, 91, 98],
    'alkphos': [55, 62, 67, 54, 60, 52, 62, 64, 77, 67, 80, 90, 72, 55, 78, 56, 57, 74, 75, 68, 47, 72, 58, 54, 50],
    'sgpt': [13, 20, 21, 22, 25, 13, 17, 61, 25, 29, 19, 43, 28, 9, 27, 30, 30, 148, 20, 17, 35, 31, 83, 25, 27],
    'sgot': [17, 17, 11, 20, 19, 24, 17, 32, 19, 20, 14, 28, 19, 25, 25, 18, 30, 75, 25, 20, 26, 20, 49, 22, 25],
    'gammagt': [17, 9, 11, 7, 5, 15, 15, 13, 18, 11, 42, 156, 30, 16, 30, 27, 22, 159, 38, 71, 33, 84, 51, 35, 53],
    'drinks': [0.0, 0.5, 0.5, 0.5, 0.5, 0.5, 0.5, 0.5, 0.5, 0.5, 2.0, 2.0, 2.0, 2.0, 2.0, 0.5, 0.5, 0.5, 0.5, 0.5, 3.0, 3.0, 3.0, 4.0, 4.0],
})

Target:
drinks

Target-specific statistics:
- Total number of values: 345
- Number of unique values: 16
- Mean value: 3.455
- Standard deviation: 3.333
- Minimum value: 0.000
- Maximum value: 20.000

\end{promptbox}
\clearpage

\section{Scaling Behavior}
\begin{figure}[H]
    \centering
    \includegraphics[width=1\textwidth]{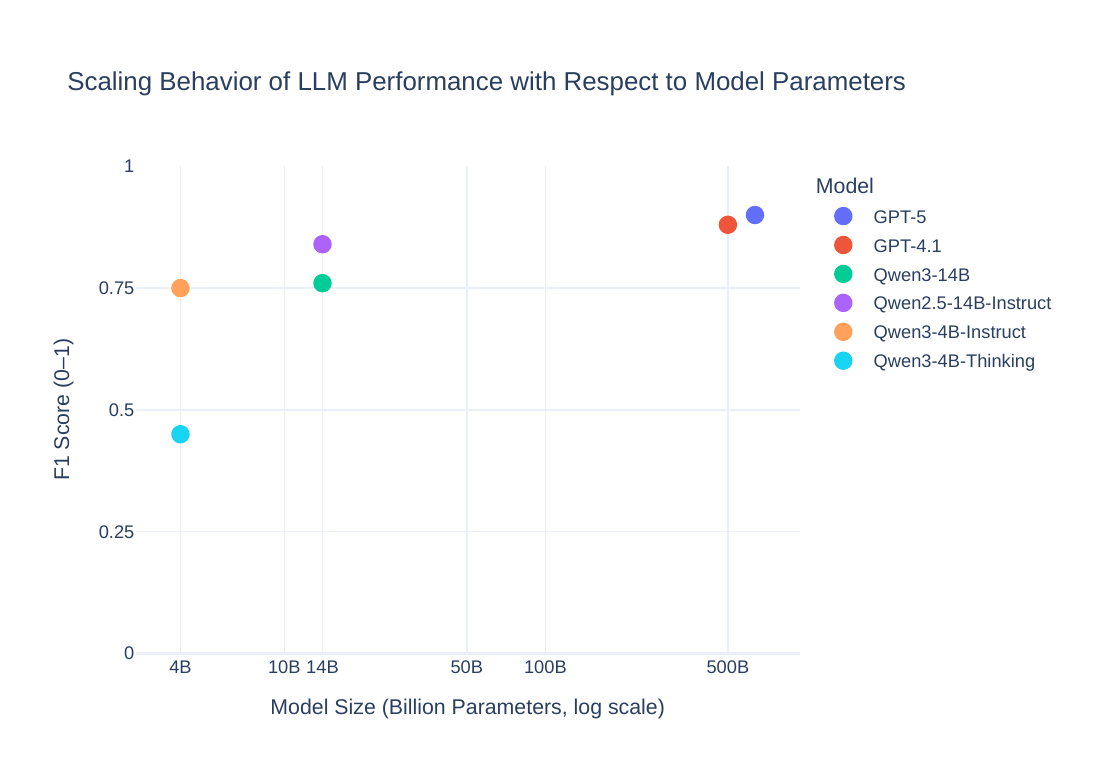}
    \caption{Scaling behavior of model performance in Dataset Phase 2.}
    \label{fig:exp3_scaling}
\end{figure}

The relationship between model size and performance is further illustrated in
Figure ~\ref{fig:exp3_scaling}. The results show a clear positive correlation between the number of model parameters and
the achieved F1 macro score. Larger models, such as GPT-4.1 and GPT-5.3, consistently outperform
smaller models, while mid-sized models such as the 14B Qwen variants achieve intermediate
performance. In contrast, smaller models such as the 4B variants show a noticeable decline in
performance, particularly in complex cross-domain settings.
This trend suggests that larger models are better able to capture both semantic and structural
aspects of the dataset, enabling more accurate task identification. As model size increases, the
ability to distinguish between subtle differences in data representation, such as tabular versus time
series structure, improves significantly. Conversely, smaller models exhibit limited capacity to
resolve these ambiguities, leading to more frequent misclassifications

\end{document}